\documentclass[runningheads]{llncs}
\usepackage[T1]{fontenc}

\usepackage{amsmath}
\usepackage{multirow}
\usepackage{hyperref}
\usepackage{graphicx}
\usepackage{amsfonts}
\usepackage{booktabs}
\usepackage{color}

\begin{document}
\title{HypeLoRA: Hyper-Network-Generated LoRA Adapters for Calibrated Language Model Fine-Tuning}
\titlerunning{HypeLoRA for Calibrated Language Model Fine-Tuning}
%
\author{Bartosz Trojan\inst{1, 2}\orcidID{0009-0005-2649-3194} \and
Filip Gębala\inst{1,3}\orcidID{0009-0002-3020-4365}}
\authorrunning{Bartosz Trojan and Filip Gębala}
%
\institute{Upper-Secondary Schools of Communications in Cracow \and
\email{bartosztrojanofficial@gmail.com} \and
\email{fgebalaofficial@gmail.com}}
\maketitle{}           
\begin{abstract}

Modern Transformer-based models frequently suffer from miscalibration, producing overconfident predictions that do not reflect true empirical frequencies. This work investigates the calibration dynamics of LoRA: Low-Rank Adaptation and a novel hyper-network-based adaptation framework as parameter-efficient alternatives to full fine-tuning for RoBERTa. Evaluating across the GLUE benchmark, we demonstrate that LoRA-based adaptation consistently achieves calibration parity with (and in specific tasks exceeds) full fine-tuning, while maintaining significantly higher parameter efficiency. We further explore a dynamic approach where a shared hyper-network generates LoRA factors (A and B matrices) to induce structural coupling across layers. This approach produced results similar to standard LoRA fine-tuning, even achieving better MCC on CoLA dataset. Our study also reveal a critical trade-off: constraining the adaptation space (e.g., freezing matrices A) acts as a powerful regularizer that enhances Expected Calibration Error (ECE), but necessitates a carefully balanced sacrifice in downstream task accuracy. To support future research, we provide a unified and reproducible implementation of contemporary calibration metrics, including ECE, MCE, and ACE. Our findings clarify the relationship between parameter efficiency and probabilistic reliability, positioning structured low-rank updates as a viable foundation for uncertainty-aware Transformer architectures. \url{https://github.com/btrojan-official/HypeLoRA}

\keywords{Fine-tuning  \and LoRA \and Hyper-network \and RoBERTa \and Calibration \and ECE}
\end{abstract}

\section{Introduction}

Transformer-based architectures have become the dominant paradigm across a wide range of machine learning domains, including natural language processing~\cite{devlin2019bert}, computer vision~\cite{dosovitskiy2020image}, and speech recognition~\cite{baevski2020wav2vec}. While these models achieve state-of-the-art predictive accuracy, it is now well established that their probabilistic outputs are often poorly calibrated~\cite{desai2020calibration}. Formally, a model is considered calibrated if the predicted confidence of a given class matches the true empirical frequency of that class among all samples assigned that confidence score — that is, among all inputs where the model outputs probability $p$, exactly a fraction $p$ should truly belong to the predicted class.



In this work, we explore parameter-efficient alternatives for calibrating transformer based encoders. We evaluate fine-tuned and augmented on LoRA RoBERTa model on the GLUE benchmark using a comprehensive set of calibration metrics, including Expected Calibration Error (ECE), Adaptive Calibration Error (ACE), and Maximum Calibration Error (MCE). In addition, we investigate the use of hyper-networks as a mechanism for inducing dynamic calibration behavior within frozen transformer architectures.

Additionally, we study a framework in which a hyper-network generates low-rank calibration signals. This setup aims to induce structural coupling across layers, forcing the model to learn a global calibration strategy rather than independent adjustments. We anticipated that this shared generation process would act as a regularizer, tempering overly sharp probability distributions. While this approach ultimately fails to yield consistent improvements, it provides valuable insight into the limitations of hyper-network-driven calibration for transformers.

Beyond empirical evaluation, we also consolidate and implement a unified set of recent calibration metrics, providing clear and reproducible reference implementations. This addresses the current fragmentation in calibration evaluation practices and facilitates more systematic comparison across methods.

Our contributions can be summarized as follows:
\begin{itemize}
\item We provide an evaluation of standard and LoRA fine-tuning for calibration of RoBERTa models on the GLUE benchmark using multiple complementary calibration metrics i.e. ECE, MCE, ACE.
\item We investigate the use of hyper-networks and LoRA as mechanisms for systematic model calibration and for analyzing the underlying causes of model failure.
\item We implement modern calibration metrics in a single evaluation framework.
\end{itemize}

\section{Related Work}

\noindent\textbf{Pre-training.} Large-scale self-supervised pre-training on generic corpora yields representations that transfer across downstream tasks, as demonstrated by masked language modeling and autoregressive objectives~\cite{devlin2019bert,liu2019roberta}. Domain- and task-adaptive pre-training on unlabeled data can further improve performance~\cite{gururangan-etal-2020-dont}, yet the high computational cost of pre-training large backbones motivates parameter-efficient adaptation methods.

\noindent\textbf{Fine-tuning large language models.} Full-parameter fine-tuning is the most direct adaptation approach, but is computationally and memory demanding for large models~\cite{lv-etal-2024-full}. When a single backbone such as RoBERTa~\cite{liu2019roberta} must be tailored to many tasks, storing full task-specific parameters and optimizer states becomes infeasible, motivating more resource-efficient alternatives.

\noindent\textbf{Parameter-efficient fine-tuning.} PEFT methods update only a small parameter subset per task while achieving performance comparable to full fine-tuning. Adapters~\cite{he2021effectiveness} insert trainable bottleneck modules into each transformer block. Prefix-tuning~\cite{li2021prefix} prepends learnable tokens to the input, optimizing a continuous prompt without modifying model weights. LoRA~\cite{hu2022lowrank} injects trainable low-rank matrices into attention layers without any activation function, substantially reducing trainable parameter count and memory cost.

\noindent\textbf{Model Calibration}~\cite{wang2023calibration}. A calibrated model's confidence estimates align with empirical likelihoods: formally, perfect calibration requires $P(Y=y|Q=q)=q$ over the joint distribution $P(Q,Y)$. We evaluate calibration primarily via expected calibration error (ECE)~\cite{10.5555/3305381.3305518}, alongside Classwise ECE, MCE, ACE, Thresholded ACE, and Brier Score~\cite{naeini2015obtaining,kull2019beyond,nixon2019measuring,Brier1950VERIFICATIONOF}. Post-hoc methods such as temperature scaling~\cite{mozafari2018attended}, Platt scaling~\cite{platt1999probabilistic}, and isotonic regression~\cite{zadrozny2002transforming} are simple but globally applied and brittle under distribution shift. Training-time approaches like label smoothing~\cite{liu2022devil} and confidence-aware regularization~\cite{liang2020neural} are more expressive but require costly retraining.

\noindent\textbf{Hyper-networks.} Hyper-networks parameterize weights as functions of external inputs, enabling dynamic generation of adaptation parameters~\cite{ha2016hyper-networks}. In PEFT, conditioning hyper-networks on task identity to generate adapter or low-rank update parameters improves parameter sharing and multi-task performance without per-task parameter duplication~\cite{ortizbarajas2024hyperloader}. Generating full weight tensors remains challenging due to scalability and stability concerns~\cite{jia2016dynamic}, so practical approaches constrain generation to compact structures such as bottleneck adapters or low-rank factors.

\section{Proposed Approach}

\begin{figure*}[t]
    \centering
    \includegraphics[width=\textwidth]{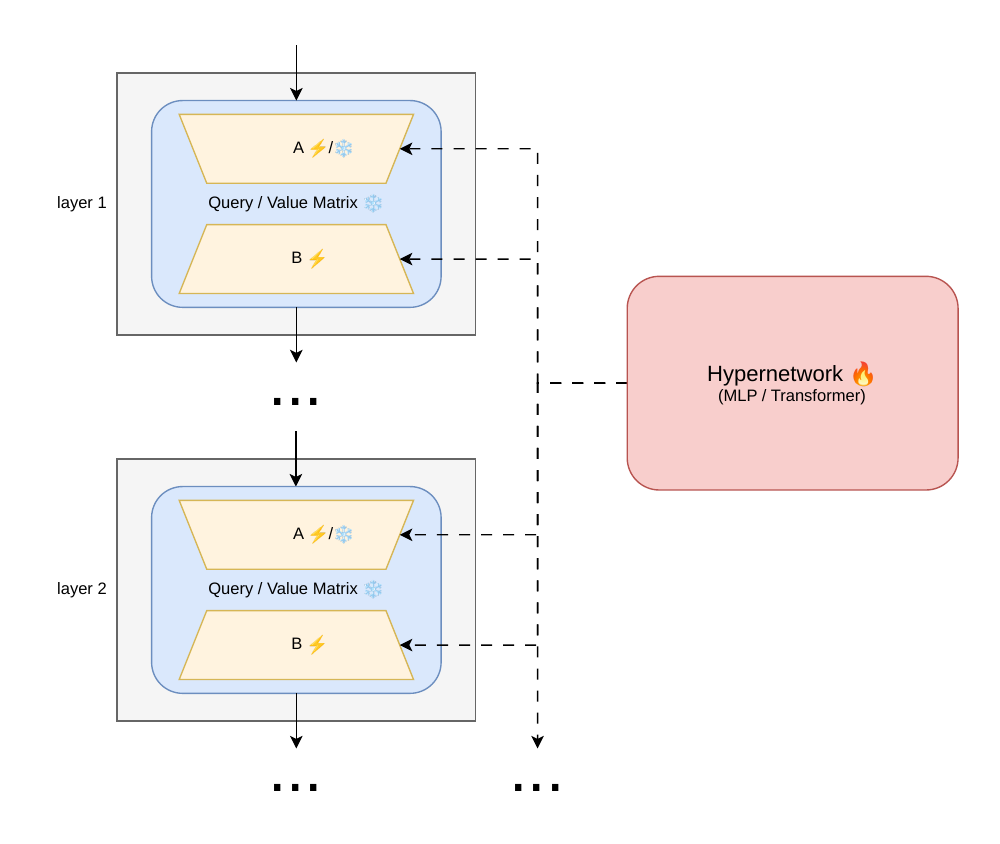}
    \caption{A hyper-network generates the weights for the Query and Value matrices in all attention blocks of the RoBERTa model, while the original pretrained weights remain frozen. The figure illustrates the approach in which the hyper-network produces both the $A$ and $B$ matrices. In this work, we also present a variant where only the $B$ matrices are generated by the hyper-network, and $A$ matrices are fixed with randomly initialized values, which isn't shown on this figure.}
    \label{fig:teaser}
\end{figure*}

We propose a parameter-efficient calibration mechanism that injects low-rank updates into a frozen RoBERTa encoder \cite{liu2019roberta} via a compact hyper-network (Figure~\ref{fig:teaser}). The hyper-network conditions on a learned embedding associated with each target weight matrix, producing coordinated adaptation signals across all transformer layers while keeping all backbone parameters frozen.

\subsection{Problem Definition}

Let $f_\theta(x)$ denote a pretrained transformer encoder with frozen parameters $\theta$, producing a probability distribution over $C$ classes for input $x$:
\begin{equation}
p_\theta(y \mid x) = \mathrm{softmax}(f_\theta(x)).
\end{equation}

Our goal is to improve predictive calibration without modifying pretrained weights. Instead of post-hoc logit adjustments (e.g., temperature scaling), we introduce structured low-rank perturbations inside the transformer layers.

\subsection{Layer-Conditioned Low-Rank Updates with Hyper-Network}

For a weight matrix $W \in \mathbb{R}^{d \times d}$ of pretrained transformer, we apply a LoRA-style \cite{hu2022lowrank} low-rank perturbation:
\begin{equation}
W' = W + \alpha\, A B,
\end{equation}
where $A \in \mathbb{R}^{d \times r}$ and $B \in \mathbb{R}^{r \times d}$ are low-rank factors of rank $r < d$, and $\alpha$ is a fixed scaling coefficient. We apply this update to the Query and Value projection matrices in each attention block.

Unlike standard LoRA, which trains all of $A$ and $B$ matrices independently, we generate these factors via a shared hyper-network $H_\phi$, conditioned on layer embedding. Each target weight matrix — specifically the Query and Value projections in each layer — is associated with a dedicated learned embedding $e \in \mathbb{R}^{d_h}$. Concretely, if both $A$s and $B$s are generated for the Query and Value matrices in layer $\ell$, there are four embeddings per layer: $e^Q_A[\ell],\, e^Q_B[\ell],\, e^V_A[\ell],\, e^V_B[\ell]$. Hyper-network $H_\phi$ maps each such embedding $e$ to target low-rank factor:
\begin{equation}
A^Q_\ell = H_\phi(e^Q_A[\ell]) \qquad B^Q_\ell = H_\phi(e^Q_B[\ell]) \qquad A^V_\ell = H_\phi(e^V_A[\ell]) \qquad B^V_\ell = H_\phi(e^V_B[\ell])
\end{equation}
The resulting vectors are reshaped into matrices of appropriate dimensions and applied to their respective weight matrices. This ties all layer adaptations through a single generator, enforcing structural coherence and drastically reducing parameter count relative to per-layer LoRA.
The architecture of $H_\phi$ is either a lightweight MLP or a small Transformer encoder operating over all embeddings jointly; implementation details are given in Section~\ref{sec:hypernet_arch}.

\textbf{Variants.} We consider two operating modes depending on whether $A$ matrices are generated or fixed:
\begin{itemize}
    \item \textbf{Full generation}: all $A$ and $B$ matrices are produced by $H_\phi$, meaning that there are four embeddings per layer in this scenario.

    \item \textbf{Fixed-A}: $A$ matrices are initialized once from a Kaiming uniform distribution (similarly to \cite{hu2022lowrank}) and held fixed throughout training; only $B$ matrices are generated by $H_\phi$, meaning that there are only two embeddings per layer. The intuition is that fixed random $A$ matrices act as structured noise injection into the frozen base model, encouraging the learned $B$ matrices to be more robust.
\end{itemize}

\subsection{Training and Inference}

During training, $H_\phi$ generates $A$s and $B$s for all target weight matrices. They are applied transiently in the forward pass — the stored pretrained weights are never modified. Gradients flow through the low-rank projections back to $H_\phi$, whose parameters $\phi$ are updated while $\theta$-original pretrained model weights-remains frozen. We optimize the standard cross-entropy loss.

At inference, the hyper-network again produces the low-rank factors on the fly, or they can be precomputed once to reduce runtime overhead. This design maintains a strict separation between the frozen backbone and the learned adaptation, ensuring a minimal memory footprint and no risk of corrupting the pretrained representations.

\section{Experimental Setup}

\subsection{Datasets}
We evaluate all of the experiments on the General Language Understanding Evaluation (GLUE) benchmark, a collection of sentence- and sentence-pair language understanding tasks designed to provide a standardized comparison of model performance across diverse NLP capabilities~\cite{wang2018glue}.
In this work, we use only GLUE tasks formulated as classification problems.

\begin{itemize}
    \item \textbf{CoLA (Corpus of Linguistic Acceptability).} A single-sentence acceptability task where the model predicts whether a sentence is linguistically acceptable (binary). CoLA has 8.5k training examples and 1k test samples.

    \item \textbf{SST-2 (Stanford Sentiment Treebank).} A single-sentence sentiment classification task where the model predicts positive vs.\ negative sentiment (binary). SST-2 has 67k training examples and 1.8k test examples.

    \item \textbf{QNLI (Question-answering Natural Language Inference).} A sentence-pair task (QA/NLI) where the model predicts whether a context sentence contains the answer to a question (binary). QNLI has 105k training examples and 5.4k test examples.

    \item \textbf{MRPC (Microsoft Research Paraphrase Corpus).} A sentence-pair paraphrase task where the model predicts whether two sentences are paraphrases (binary). MRPC has 3.7k training examples and 1.7k test examples.

    \item \textbf{RTE (Recognizing Textual Entailment).} A sentence-pair inference task (NLI) where the model predicts whether a hypothesis is entailed by a premise (binary). RTE has 2.5k training and 3k test examples.

    \item \textbf{MNLI (Multi-Genre Natural Language Inference).} A sentence-pair natural language inference task where the model predicts the relationship between a premise and a hypothesis (entailment, contradiction, or neutral; three-way classification). MNLI has 393k training examples and 20k test examples.

\end{itemize}

\subsection{Training and Evaluation Setup}
Unless stated otherwise, we follow the experimental protocol from the original LoRA work~\cite{hu2022lowrank} for training and evaluation.
The two main differences are (i) the introduction of a hyper-network to generate the low-rank update parameters and (ii) the specific injection strategy used to apply these generated updates within the frozen RoBERTa encoder \cite{liu2019roberta}. For our approach, we use different peak learning rates (e.g., $1\mathrm{e}{-5}$ for the MLP and $4\mathrm{e}{-4}$ for the Transformer), as we observed that lower learning rates lead to more stable and effective training when using the MLP-based hyper-network.

\subsection{Hyper-network Architectures}
\label{sec:hypernet_arch}

We consider two hyper-network architectures for generating the LoRA factors: a multilayer perceptron (MLP) and a Transformer encoder \cite{vaswani2017attention}. In both cases, each transformer layer identifier $\ell$ is represented by a learnable embedding of dimension 128.

\textbf{MLP.}
The MLP consists of four fully connected layers. The input dimension is 128, all hidden layers have width 2048, and the GELU as its activation function. The output layer projects to the flattened LoRA parameter space. In all experiments, the hidden size of RoBERTa is 768 and the LoRA rank is $r = 8$, so the output dimension corresponds to $768 \times 8$ per generated factor.

All MLP weights are initialized from a normal distribution. In the configuration where $A$ matrices are fixed, it is initialized once using Kaiming uniform initialization and kept frozen, while only $B$ is generated by the hyper-network.

\textbf{Transformer.}
In the Transformer-based variant, all layer embeddings (dimension 128) are first projected with a linear layer to a hidden dimension of 256 and then processed jointly by a Transformer encoder with 2 layers, 16 attention heads, and model dimension 256. The outputs corresponding to each layer are passed through a final linear projection to produce the flattened LoRA parameters, either for both $A$ and $B$ matrices or only for $B$s when $A$s are fixed.

All learnable parameters are initialized from a normal distribution, except for the fixed $A$ matrices, which use Kaiming uniform initialization.

\subsection{Metrics}
\label{sec:metrics}

Exact calibration measurement with finite samples is not possible due to continuous predicted confidence values. In practice, calibration error is approximated by partitioning $N$ predictions into $M$ bins $\{b_1, \dots, b_M\}$ based on predicted probabilities and comparing average confidence with empirical accuracy within each bin. $M = 10$ for all of our experiments.

The most widely used metric is \textbf{Expected Calibration Error (ECE)} \cite{naeini2015obtaining}, defined as the weighted average of per-bin confidence--accuracy gaps:
\begin{equation}
\mathrm{ECE} = \sum_{m=1}^{M} \frac{|b_m|}{N} \left| \mathrm{acc}(b_m) - \mathrm{conf}(b_m) \right|,
\end{equation}
where $|b_m|$ is the number of samples in bin $b_m$, $\mathrm{acc}(b_m)$ is the fraction of correct predictions, and $\mathrm{conf}(b_m)$ is the mean predicted probability within that bin.

The remaining metrics follow the same bin-level discrepancy idea with specific modifications. \textbf{Maximum Calibration Error (MCE)} \cite{naeini2015obtaining} replaces the weighted average with the worst-case bin maximum. \textbf{Classwise ECE (CECE)} \cite{kull2019beyond} computes the ECE-style sum separately for each class and averages across all $K$ classes. \textbf{Adaptive Calibration Error (ACE)} \cite{nixon2019measuring} replaces fixed-width bins with equal-population bins, also averaging per class. \textbf{Thresholded ACE (TACE)} \cite{nixon2019measuring} further restricts ACE to predictions whose confidence exceeds a threshold $\epsilon$, reducing the influence of near-zero probabilities. Finally, the \textbf{Brier Score (BS)} \cite{Brier1950VERIFICATIONOF} is a proper scoring rule computing the mean squared error between predicted probabilities and one-hot targets across all classes, providing a combined measure of calibration and sharpness.

\section{Results}

Here we present the results of our experiments, including calibration analysis for standard Fine-Tuning (FT), Low-Rank Adaptation (LoRA), and our proposed hyper-network-based variants.

\subsection{Calibration of the Existing Methods}

\begin{table}[t]
\caption{Performance and calibration metrics across GLUE benchmarks for RoBERTa$_{\text{large}}$. FT denotes standard fine-tuning \cite{liu2019roberta}; LoRA denotes Low-Rank Adaptation \cite{hu2022lowrank}. Following \cite{wang2018glue}, we report Matthews Correlation Coefficient (MCC) for CoLA, F1 for MRPC, and Accuracy for other tasks. Minor score deviations from \cite{hu2022lowrank} are attributable to different random seeds; additionally, MRPC reports F1 (as in \cite{liu2019roberta}) rather than accuracy, and RTE (FT) is initialized from MNLI-pretrained weights for fair comparison \cite{hu2022lowrank}.}
\label{tab1}
\centering
\scriptsize
\begin{tabular}{ l c c c c c c c}
\toprule
 & Metric & CoLA & SST-2 & QNLI & MRPC & RTE & MNLI \\
\midrule
\multirow{8}{*}{FT} & $Score$ $\uparrow$ & 61.68 $\pm_{0.73}$ & 94.50 $\pm_{0.41}$ & 92.64 $\pm_{0.22}$ & 90.37 $\pm_{0.41}$ & 85.32 $\pm_{1.37}$ & 87.17  \\
 & ECE $\downarrow$ & 0.136 $\pm_{0.008}$ & 0.035 $\pm_{0.017}$ & 0.072 $\pm_{0.002}$ & 0.111 $\pm_{0.023}$ & 0.104 $\pm_{0.045}$ & 0.074  \\
 & CECE $\downarrow$ & 0.138 $\pm_{0.007}$ & 0.036 $\pm_{0.016}$ & 0.072 $\pm_{0.002}$ & 0.114 $\pm_{0.021}$ & 0.113 $\pm_{0.038}$ & 0.050  \\
  & MCE $\downarrow$ & 0.339 $\pm_{0.074}$ & 0.277 $\pm_{0.242}$ & 0.539 $\pm_{0.111}$ & 0.308 $\pm_{0.163}$ & 0.309 $\pm_{0.159}$ & 0.202  \\
 & ACE $\downarrow$ & 0.134 $\pm_{0.009}$ & 0.033 $\pm_{0.016}$ & 0.071 $\pm_{0.002}$ & 0.110 $\pm_{0.021}$ & 0.103 $\pm_{0.039}$ & 0.073  \\
 & TACE$_{0.01}$ $\downarrow$ & 0.469 $\pm_{0.006}$ & 0.461 $\pm_{0.013}$ & 0.482 $\pm_{0.007}$ & 0.441 $\pm_{0.020}$ & 0.413 $\pm_{0.058}$ & 0.492  \\
 & Brier Score $\downarrow$ & 0.290 $\pm_{0.006}$ & 0.096 $\pm_{0.006}$ & 0.145 $\pm_{0.004}$ & 0.245 $\pm_{0.013}$ & 0.266 $\pm_{0.009}$ & 0.207  \\
\midrule
\multirow{8}{*}{LoRA} & $Score$ $\uparrow$ & 63.94 $\pm_{0.21}$ & 94.99 $\pm_{0.18}$ & 93.07 $\pm_{0.05}$ & 93.18 $\pm_{0.40}$ & 87.61 $\pm_{0.21}$ & 87.19  \\
 & ECE $\downarrow$ & 0.120 $\pm_{0.025}$ & 0.046 $\pm_{0.001}$ & 0.036 $\pm_{0.011}$ & 0.088 $\pm_{0.012}$ & 0.124 $\pm_{0.007}$ & 0.043  \\
 & CECE $\downarrow$ & 0.123 $\pm_{0.024}$ & 0.046 $\pm_{0.001}$ & 0.037 $\pm_{0.011}$ & 0.088 $\pm_{0.011}$ & 0.125 $\pm_{0.006}$ & 0.029  \\
  & MCE $\downarrow$ & 0.282 $\pm_{0.038}$ & 0.340 $\pm_{0.111}$ & 0.123 $\pm_{0.052}$ & 0.424 $\pm_{0.097}$ & 0.502 $\pm_{0.156}$ & 0.257  \\
 & ACE $\downarrow$ & 0.114 $\pm_{0.024}$ & 0.040 $\pm_{0.003}$ & 0.035 $\pm_{0.010}$ & 0.084 $\pm_{0.012}$ & 0.114 $\pm_{0.002}$ & 0.043  \\
 & TACE$_{0.01}$ $\downarrow$ & 0.443 $\pm_{0.026}$ & 0.477 $\pm_{0.007}$ & 0.452 $\pm_{0.012}$ & 0.447 $\pm_{0.005}$ & 0.451 $\pm_{0.024}$ & 0.432  \\
 & Brier Score $\downarrow$ & 0.271 $\pm_{0.020}$ & 0.096 $\pm_{0.002}$ & 0.114 $\pm_{0.004}$ & 0.180 $\pm_{0.017}$ & 0.244 $\pm_{0.009}$ & 0.193  \\
\bottomrule
\end{tabular}
\end{table}

Table~\ref{tab1} presents task performance and calibration metrics for standard Fine-Tuning and LoRA fine-tuning across selected GLUE benchmarks. LoRA consistently matches or improves predictive performance relative to full Fine-Tuning, with noticeable gains on CoLA, MRPC, QNLI, and RTE, while maintaining comparable results on MNLI.

From a calibration perspective, ECE remains relatively stable across seeds for both methods, typically exhibiting low variance. CECE closely follows ECE, reflecting the predominantly binary structure of the evaluated tasks. As expected, MCE shows substantially higher variability due to its sensitivity to worst-case confidence bins. ACE aligns closely with ECE, indicating an approximately uniform distribution of samples across confidence bins.

Across tasks, LoRA demonstrates improved calibration on QNLI and MNLI, where ECE and Brier Score are consistently lower than under full Fine-Tuning. However, this behavior is not uniform. On SST-2 and RTE, Fine-Tuning achieves lower ECE values, suggesting better calibration despite slightly weaker predictive performance. The elevated values of $\text{TACE}_{0.01}$ (TACE with threshold equal to $0.01$) across both methods indicate that predictions are concentrated near extreme probabilities, leading to high-confidence outputs and increased calibration penalties when misclassifications occur.

Overall, LoRA does not uniformly improve calibration over full Fine-Tuning. Its effect is task-dependent, and improvements in predictive performance do not systematically translate into better uncertainty estimation.

\subsection{Our Approach - Calibration with Hyper-Network}

\begin{table}[ht]
\centering
\caption{We compare four hypernetwork configurations combining two architectures (MLP and Transformer) with two weight generation strategies ($A_{\text{gen}}$ and $A_{\text{fix}}$), each averaged over 3 seeds. Transformer $A_{\text{gen}}$ outperforms the LoRA baseline on CoLA, though LoRA remains stronger on SST-2, suggesting task-dependent behavior. Interestingly, Transformer $A_{\text{fix}}$ achieves the best calibration across all runs despite not leading in Matthews Correlation Coefficient (MCC) or Accuracy.}
\label{tab:results}
\begin{tabular}{lcccc}
\toprule
 & \multicolumn{2}{c}{COLA} & \multicolumn{2}{c}{SST-2} \\
 & MCC $\uparrow$ & ECE $\downarrow$ & Acc $\uparrow$ & ECE $\downarrow$ \\
\midrule
MLP $A_\text{gen}$ & $63.54_{\pm0.31}$ & $0.116_{\pm0.019}$ & $94.84_{\pm0.30}$ & $0.037_{\pm0.007}$ \\
MLP $A_\text{fix}$ & $48.25_{\pm13.01}$ & $0.130_{\pm0.015}$ & $92.89_{\pm1.41}$ & $0.047_{\pm0.011}$ \\
Transformer $A_\text{gen}$ & $\mathbf{64.42_{\pm1.75}}$ & $0.119_{\pm0.009}$ & $94.78_{\pm0.08}$ & $0.040_{\pm0.003}$ \\
Transformer $A_\text{fix}$ & $60.69_{\pm0.35}$ & $\mathbf{0.100_{\pm0.010}}$ & $94.56_{\pm0.08}$ & $\mathbf{0.028_{\pm0.004}}$ \\
\midrule
LoRA & $63.94_{\pm0.21}$ & $0.120_{\pm0.025}$ & $\mathbf{94.99_{\pm0.18}}$ & $0.046_{\pm0.001}$ \\
\bottomrule
\end{tabular}
\end{table}

\begin{figure*}[t]
    \centering
    \includegraphics[width=\textwidth]{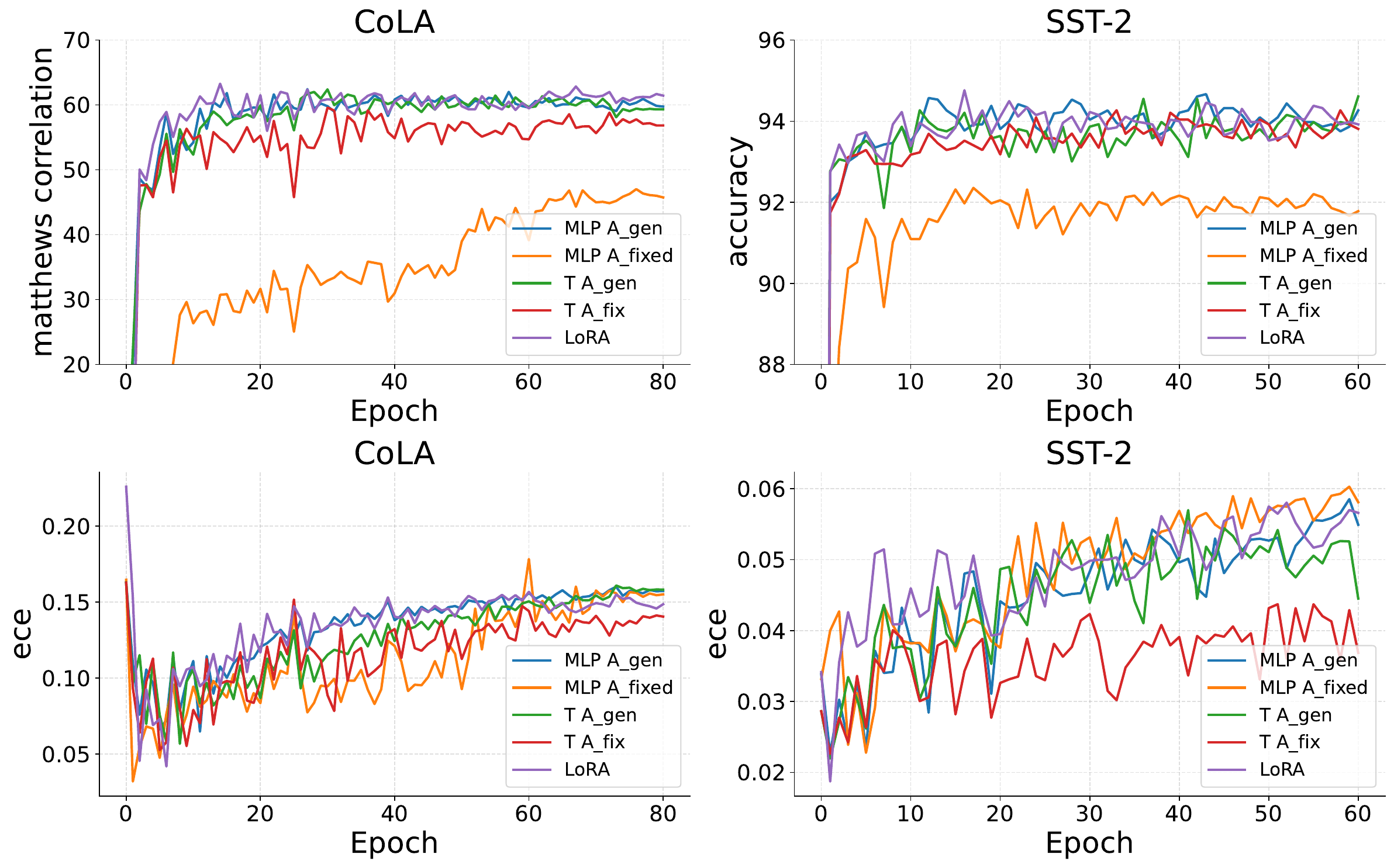}
    \caption{Evaluation results on CoLA and SST-2 benchmarks, reported as Matthews Correlation Coefficient and accuracy (top row) alongside Expected Calibration Error (bottom row), averaged across 3 independent random seeds. $A_\text{gen}$ means both matrices $A$ and $B$ are generated, and $A_\text{fix}$ means matrices $A$ are fixed. LoRA~\cite{hu2022lowrank} is included as a baseline. Fixing matrix $A$ improves model calibration, albeit at the cost of task performance across both datasets.}
    \label{fig:all_charts}
\end{figure*}

Table~\ref{tab:results} reports the performance and calibration of the proposed hyper-network variants on CoLA and SST-2, compared to the LoRA baseline. We evaluate both MLP-based and Transformer-based hyper-networks, with either full generation of adapter matrices ($A_{\text{gen}}$) or with matrices $A$ fixed ($A_{\text{fix}}$).

On CoLA, the Transformer-based hyper-network with fully generated matrices (Transformer $A_{\text{gen}}$) achieves average highest performance, but also shows high variability across different seeds. It surpasses LoRA fine-tuning. However, its calibration remains similar to LoRA, indicating no inherent calibration advantage from full matrix generation. When $A$ matrices are fixed (Transformer $A_{\text{fix}}$), MCC value decreases, but ECE improves, representing the best calibration among all evaluated configurations. We hypothesize that this calibration improvement is caused by a reduction in effective degrees of freedom - fixing $A$ matrices constrains the adaptation space, acting as an implicit regularizer that prevents the model from becoming overconfident.

The MLP-based variants follow the same qualitative pattern but exhibit reduced stability. While MLP $A_{\text{gen}}$ maintains competitive performance, fixing $A$ matrices leads to a substantial drop in performance without improving calibration relative to the Transformer-based fixed configuration.

On SST-2, LoRA achieves the highest predictive accuracy. Both hyper-network variants with generated matrices remain competitive. Again, freezing matrices $A$ results in a small but consistent decrease in accuracy. However, the Transformer-based fixed configuration yields the lowest ECE, significantly outperforming both LoRA and the fully generated variant.

Contrary to our initial hypothesis, fully generated hyper-network adapters do not systematically improve calibration over LoRA, despite occasionally improving task performance (notably on CoLA). Instead, calibration improvements emerge primarily when matrices $A$ are fixed. This constraint introduces a structured transformation of the adapter input space, limiting overconfident predictions and acting as a form of implicit regularization. The improvement in calibration is accompanied by a clear trade-off in predictive performance, particularly pronounced on CoLA.

Additionally, as show on Fig. \ref{fig:all_charts}, we observe that extended training consistently leads to worsening calibration across all configurations, even when task metrics continue to improve. This behavior suggests progressive overfitting to the training objective, resulting in sharper predictive distributions and degraded uncertainty estimation.

\section{Conclusion}
In this work, we investigated parameter-efficient adaptation mechanisms as a novel approach to calibration for transformer-based language models. We analyzed standard Fine-Tuning and LoRA fine-tuning \cite{hu2022lowrank} applied to RoBERTa \cite{liu2019roberta}, and introduced a hyper-network-based variant in which low-rank update matrices are generated conditionally on transformer layer identity.

Our evaluation across multiple GLUE benchmarks \cite{wang2018glue} yields three main findings. First, LoRA provides calibration comparable to full Fine-Tuning while retaining parameter efficiency, though calibration improvements remain task-dependent. Second, hyper-network-generated low-rank factors yield calibration broadly similar to standard LoRA, suggesting structural cross-layer coupling alone is insufficient for systematic confidence correction, but also our proposed Transformer-based hyper-network LoRA variant (Transformer $A_{\text{gen}}$) showed promising results, by outperforming standard LoRA on the CoLA benchmark. Third, freezing all of the $A$ matrices ($A_{\text{fix}}$) modestly improves calibration at the cost of task performance, revealing a tension between representation stability and predictive sharpness.
We additionally provide a unified, reproducible implementation of six calibration metrics (ECE, CECE, MCE, ACE, TACE, Brier Score), contributing toward more systematic evaluation standards in transformer calibration research.

\textbf{Further Work} Future work should investigate the mechanism behind the $A_{\text{fix}}$ calibration improvement — whether it stems from reduced flexibility, additional noise, or modified optimization dynamics. The CoLA advantage of the Transformer hyper-network motivates further study of this architecture. Extending evaluation to multi-class and out-of-distribution benchmarks would clarify whether observed improvements generalize beyond binary GLUE tasks.

\begin{credits}
\subsubsection{\ackname}
The authors are grateful to Dr. Kamil Książek, Dr. Tomasz Kuśmierczyk, and Prof. Jacek Tabor of the Jagiellonian University for their invaluable guidance and for  providing access to computational resources.
\subsubsection{\discintname}
The authors have no competing interests to declare that are relevant to the content of this article.
\end{credits}
%
%
%
%

\bibliographystyle{splncs04}
\bibliography{bibliography}

@inproceedings{hu2022lowrank,
  author = {Hu, Edward J. and Shen, Yelong and Wallis, Phillip and Allen-Zhu, Zeyuan and Li, Yuanzhi and Wang, Shean and Wang, Lu and Chen, Weizhu},
  booktitle = {ICLR},
  ee = {https://openreview.net/forum?id=nZeVKeeFYf9},
  publisher = {OpenReview.net},
  title = {LoRA: Low-Rank Adaptation of Large Language Models.},
  url = {http://dblp.uni-trier.de/db/conf/iclr/iclr2022.html\#HuSWALWWC22},
  year = 2022
}

@inproceedings{liu2019roberta,
  title={RoBERTa: A Robustly Optimized BERT Pretraining Approach},
  author={Liu, Yinhan and Ott, Myle and Goyal, Naman and Du, Jingfei and Joshi, Mandar and Chen, Danqi and Levy, Omer and Lewis, Mike and Zettlemoyer, Luke and Stoyanov, Veselin},
  booktitle={International Conference on Learning Representations (ICLR)},
  year={2020}
}

@article{desai2020calibration,
  title={Calibration of pre-trained transformers},
  author={Desai, Shrey and Durrett, Greg},
  journal={arXiv preprint arXiv:2003.07892},
  year={2020}
}

@inproceedings{wang2018glue,
  title={GLUE: A multi-task benchmark and analysis platform for natural language understanding},
  author={Wang, Alex and Singh, Amanpreet and Michael, Julian and Hill, Felix and Levy, Omer and Bowman, Samuel},
  booktitle={Proceedings of the 2018 EMNLP workshop BlackboxNLP: Analyzing and interpreting neural networks for NLP},
  pages={353--355},
  year={2018}
}

@inproceedings{lv-etal-2024-full,
    title = "Full Parameter Fine-tuning for Large Language Models with Limited Resources",
    author = "Lv, Kai  and
      Yang, Yuqing  and
      Liu, Tengxiao  and
      Guo, Qipeng  and
      Qiu, Xipeng",
    editor = "Ku, Lun-Wei  and
      Martins, Andre  and
      Srikumar, Vivek",
    booktitle = "Proceedings of the 62nd Annual Meeting of the Association for Computational Linguistics (Volume 1: Long Papers)",
    month = aug,
    year = "2024",
    address = "Bangkok, Thailand",
    publisher = "Association for Computational Linguistics",
    url = "https://aclanthology.org/2024.acl-long.445/",
    doi = "10.18653/v1/2024.acl-long.445",
    pages = "8187--8198"
}

@article{li2021prefix,
  title={Prefix-tuning: Optimizing continuous prompts for generation},
  author={Li, Xiang Lisa and Liang, Percy},
  journal={arXiv preprint arXiv:2101.00190},
  year={2021}
}

@inproceedings{devlin2019bert,
  title={Bert: Pre-training of deep bidirectional transformers for language understanding},
  author={Devlin, Jacob and Chang, Ming-Wei and Lee, Kenton and Toutanova, Kristina},
  booktitle={Proceedings of the 2019 conference of the North American chapter of the association for computational linguistics: human language technologies, volume 1 (long and short papers)},
  pages={4171--4186},
  year={2019}
}

@misc{ortizbarajas2024hyperloader,
  title={HyperLoader: Integrating Hypernetwork-Based LoRA and Adapter Layers into Multi-Task Transformers for Sequence Labelling},
  author={Ortiz-Barajas, Jes{\'u}s-Germ{\'a}n and G{\'o}mez-Adorno, Helena and Solorio, Thamar},
  year={2024},
  eprint={2407.01411},
  archivePrefix={arXiv},
  primaryClass={cs.CL}
}

@inproceedings{gururangan-etal-2020-dont,
    title = "Don{'}t Stop Pretraining: Adapt Language Models to Domains and Tasks",
    author = "Gururangan, Suchin  and
      Marasovi{\'c}, Ana  and
      Swayamdipta, Swabha  and
      Lo, Kyle  and
      Beltagy, Iz  and
      Downey, Doug  and
      Smith, Noah A.",
    editor = "Jurafsky, Dan  and
      Chai, Joyce  and
      Schluter, Natalie  and
      Tetreault, Joel",
    booktitle = "Proceedings of the 58th Annual Meeting of the Association for Computational Linguistics",
    month = jul,
    year = "2020",
    address = "Online",
    publisher = "Association for Computational Linguistics",
    url = "https://aclanthology.org/2020.acl-main.740/",
    doi = "10.18653/v1/2020.acl-main.740",
    pages = "8342--8360"
}

@article{dosovitskiy2020image,
  title={An image is worth 16x16 words: Transformers for image recognition at scale},
  author={Dosovitskiy, Alexey},
  journal={arXiv preprint arXiv:2010.11929},
  year={2020}
}

@article{baevski2020wav2vec,
  title={wav2vec 2.0: A framework for self-supervised learning of speech representations},
  author={Baevski, Alexei and Zhou, Yuhao and Mohamed, Abdelrahman and Auli, Michael},
  journal={Advances in neural information processing systems},
  volume={33},
  pages={12449--12460},
  year={2020}
}

@inproceedings{10.5555/3305381.3305518,
author = {Guo, Chuan and Pleiss, Geoff and Sun, Yu and Weinberger, Kilian Q.},
title = {On calibration of modern neural networks},
year = {2017},
publisher = {JMLR.org},
booktitle = {Proceedings of the 34th International Conference on Machine Learning - Volume 70},
pages = {1321–1330},
numpages = {10},
location = {Sydney, NSW, Australia},
series = {ICML'17}
}

@inproceedings{naeini2015obtaining,
  title={Obtaining well calibrated probabilities using bayesian binning},
  author={Naeini, Mahdi Pakdaman and Cooper, Gregory and Hauskrecht, Milos},
  booktitle={Proceedings of the AAAI conference on artificial intelligence},
  volume={29},
  number={1},
  year={2015}
}

@article{kull2019beyond,
  title={Beyond temperature scaling: Obtaining well-calibrated multi-class probabilities with dirichlet calibration},
  author={Kull, Meelis and Perello Nieto, Miquel and K{\"a}ngsepp, Markus and Silva Filho, Telmo and Song, Hao and Flach, Peter},
  journal={Advances in neural information processing systems},
  volume={32},
  year={2019}
}

@inproceedings{nixon2019measuring,
  title={Measuring calibration in deep learning.},
  author={Nixon, Jeremy and Dusenberry, Michael W and Zhang, Linchuan and Jerfel, Ghassen and Tran, Dustin},
  booktitle={CVPR workshops},
  volume={2},
  number={7},
  year={2019}
}

@article{Brier1950VERIFICATIONOF,
  title={VERIFICATION OF FORECASTS EXPRESSED IN TERMS OF PROBABILITY},
  author={Glenn W. Brier},
  journal={Monthly Weather Review},
  year={1950},
  volume={78},
  pages={1-3},
  url={https://api.semanticscholar.org/CorpusID:122906757}
}

@article{vaswani2017attention,
  title={Attention is all you need},
  author={Vaswani, Ashish and Shazeer, Noam and Parmar, Niki and Uszkoreit, Jakob and Jones, Llion and Gomez, Aidan N and Kaiser, {\L}ukasz and Polosukhin, Illia},
  journal={Advances in neural information processing systems},
  volume={30},
  year={2017}
}

@article{wang2023calibration,
  title={Calibration in deep learning: A survey of the state-of-the-art},
  author={Wang, Cheng},
  journal={arXiv preprint arXiv:2308.01222},
  year={2023}
}

@article{mozafari2018attended,
  title={Attended temperature scaling: a practical approach for calibrating deep neural networks},
  author={Mozafari, Azadeh Sadat and Gomes, Hugo Siqueira and Le{\~a}o, Wilson and Janny, Steeven and Gagn{\'e}, Christian},
  journal={arXiv preprint arXiv:1810.11586},
  year={2018}
}

@article{platt1999probabilistic,
  title={Probabilistic outputs for support vector machines and comparisons to regularized likelihood methods},
  author={Platt, John and others},
  journal={Advances in large margin classifiers},
  volume={10},
  number={3},
  pages={61--74},
  year={1999},
  publisher={Cambridge, MA}
}

@inproceedings{zadrozny2002transforming,
  title={Transforming classifier scores into accurate multiclass probability estimates},
  author={Zadrozny, Bianca and Elkan, Charles},
  booktitle={Proceedings of the eighth ACM SIGKDD international conference on Knowledge discovery and data mining},
  pages={694--699},
  year={2002}
}

@inproceedings{liu2022devil,
  title={The devil is in the margin: Margin-based label smoothing for network calibration},
  author={Liu, Bingyuan and Ben Ayed, Ismail and Galdran, Adrian and Dolz, Jose},
  booktitle={Proceedings of the IEEE/CVF Conference on Computer Vision and Pattern Recognition},
  pages={80--88},
  year={2022}
}

@inproceedings{liang2020neural,
  title={Neural network calibration for medical imaging classification using dca regularization},
  author={Liang, Gongbo and Zhang, Yu and Jacobs, Nathan},
  booktitle={International conference on machine learning, workshop on uncertainty and robustness in deep learning},
  year={2020}
}

@article{jia2016dynamic,
  title={Dynamic filter networks},
  author={Jia, Xu and De Brabandere, Bert and Tuytelaars, Tinne and Gool, Luc V},
  journal={Advances in neural information processing systems},
  volume={29},
  year={2016}
}

@article{ha2016hyper-networks,
  title={Hypernetworks},
  author={Ha, David and Dai, Andrew and Le, Quoc V},
  journal={arXiv preprint arXiv:1609.09106},
  year={2016}
}

@inproceedings{he2021effectiveness,
  title={On the effectiveness of adapter-based tuning for pretrained language model adaptation},
  author={He, Ruidan and Liu, Linlin and Ye, Hai and Tan, Qingyu and Ding, Bosheng and Cheng, Liying and Low, Jiawei and Bing, Lidong and Si, Luo},
  booktitle={Proceedings of the 59th annual meeting of the association for computational linguistics and the 11th international joint conference on natural language processing (volume 1: long papers)},
  pages={2208--2222},
  year={2021}
}

\end{document}